# Generating Counterfactual Patient Timelines from Real-World Data

Yu Akagi, M.D.[1], Tomohisa Seki, M.D., Ph.D.[2],
Toru Takiguchi, M.D., Ph.D.[2], Hiromasa Ito, M.D., Ph.D.[2],
Yoshimasa Kawazoe, M.D., Ph.D.[2,3], Kazuhiko Ohe, M.D., Ph.D.[1,2]

[1] Department of Biomedical Informatics, Graduate School of Medicine,
The University of Tokyo, Japan
[2] Department of Healthcare Information Management, The University of Tokyo Hospital,
Tokyo, Japan
[3] Artificial Intelligence and Digital Twin in Healthcare, Graduate School of Medicine, The
University of Tokyo, Tokyo, Japan

**Abstract**
*Counterfactual simulation—exploring hypothetical consequences under alternative clinical scenarios—holds promise for transformative applications such as personalized medicine and in-silico trials. However, it remains challenging due to methodological limitations. Here, we show that an autoregressive generative model, trained on real-world data from over 300,000 patients and 400 million patient timeline entries, can generate clinically plausible counterfactual trajectories. As a validation task, we applied the model to patients hospitalized with COVID-19 in 2023, modifying age, serum C-reactive protein (CRP), and serum creatinine to simulate 7-day outcomes. Increased in-hospital mortality was observed in counterfactual simulations with older age, elevated CRP, and elevated serum creatinine. Remdesivir prescriptions increased in simulations with higher CRP values and decreased in those with impaired kidney function. These counterfactual trajectories reproduced known clinical patterns. These findings suggest that autoregressive generative models trained on real-world data in a self-supervised manner can establish a foundation for counterfactual clinical simulation.*

**Introduction**

Counterfactual simulation—simulation of "what-if" scenarios—holds promise for a wide range of transformative clinical applications. In medicine, where physicians routinely choose among multiple treatment options, the ability to forecast alternative outcomes could support personalized care and enhance therapeutic decision-making[1,2]. If accurate, counterfactual models may supplement clinical trials by enabling in-silico comparisons of interventions[3]. Yet, despite this promise, counterfactual simulation at the patient level remains infeasible. A central challenge lies in the simulation itself: patient trajectories emerge from complex physiological responses and clinician-driven treatment decisions, both of which are difficult to capture using traditional modeling approaches. Developing models that can learn and reproduce these dynamics is essential for general-purpose counterfactual simulation[4].

Recent advances in generative artificial intelligence offer promising solutions to these challenges[5]. In particular, autoregressive decoding of patient timelines has emerged as a method for generating patient trajectories conditioned on existing health records[5–7]. While the specific architectures and data formats vary across studies, these models fundamentally operate in a manner analogous to large language models. They sequentially generate clinical entities such as diagnoses, medications, or laboratory results instead of words[8,9]. Although previous work has demonstrated that such models can produce apparently realistic patient timelines, their potential for counterfactual simulation remains unexplored.

In this study, we developed a transformer-based autoregressive model trained on over 400 million timeline entries from more than 300,000 patients. We applied the model to simulate 7-day outcomes for patients hospitalized with COVID-19 in 2023. To test counterfactual reasoning, we modified the key input features: age, serum C-reactive protein (CRP), and serum creatinine. We then examined how simulated outcomes changed in response to these modifications. The resulting trajectories aligned with established clinical knowledge, such as increased mortality with increased serum CRP and reduced remdesivir use with impaired kidney function. These findings demonstrate that autoregressive decoding models can serve as patient-level counterfactual simulators, opening new possibilities for personalized risk assessment, simulation-based research, and decision support in real-world care.

**Methods**

This study was reviewed and approved by the Research Ethics Committee of the University of Tokyo (Approval Number: 2023367NI). All procedures were conducted in accordance with relevant ethical guidelines and institutional regulations.

**Data acquisition**

We accessed longitudinal electronic health records from the University of Tokyo Hospital, spanning January 2011 to December 2023. The dataset comprises patients of all ages and covers all hospital departments, including admissions and outpatient visits. Structured data were extracted for patient demographics (sex only), admission and discharge events, medication prescriptions, and laboratory test results. All records were obtained via Health Level Seven (HL7) version 2.5 standardized messaging. Diagnosis and medication codes were mapped to the International Classification of Diseases, 10th Revision (ICD-10), and the Anatomical Therapeutic Chemical (ATC) classification system, respectively. Laboratory test identifiers followed the Japanese JLAC10 coding standard without further transformation.

**Mathematical formulation of the autoregressive decoding simulation model**

The model's data generation process is illustrated in Figure 1. The simulation model developed in this study is an autoregressive deep-learning model built on a decoder-only transformer backbone. It generates patient timelines by predicting one entry at a time, conditioned on previously generated entries.

Let the model operate over a finite vocabulary:

$$\Omega = \{\omega_1, \omega_3, \dots, \omega_V\}$$

, where $V$ is the vocabulary size and each $\omega$ represents a unique timeline entry. Given a patient timeline of length $t$, we denote the timeline as:

$$S_t = (e_1, e_2, \dots, e_t) \in \Omega^t$$

, where each $e_i$ is a timeline entry. The model is defined as a function $f$ that maps the sequence $S_t$ to a vector of logits over the vocabulary:

$$f: \Omega^t \to \mathbb{R}^V$$

. At each timestep $t$, the model produces a logit vector:

$$\boldsymbol{z}_{t+1} = f(S_t; \theta)$$

, where $\theta$ is the trainable parameters of the model. To obtain the probability distribution over the vocabulary for the next entry at position $t + 1$, we apply the softmax function:

$$\boldsymbol{p}_{t+1} = \text{softmax}(\boldsymbol{z}_t) = \frac{\exp(\boldsymbol{z}_t)}{\sum_{i=1}^{V} \exp(\boldsymbol{z}_{t,i})}$$

The model samples the next entry from the probability distribution:

$$e_{t+1} \sim \boldsymbol{p}_{t+1}$$

. The model repeats this process autoregressively, appending each newly generated entry $e_{t+1}$ to the input sequence $S_t$, forming an updated sequence $S_{t+1}$:

$$S_{t+1} = (e_1, e_2, \dots, e_t, e_{t+1})$$

This process continues until a stopping condition is met, such as reaching a specified time horizon or generating a special end-of-trajectory entry. Since the model generates each entry conditioned on previously generated entries, the probability of producing an entire patient timeline $S_T$ is factorized as:

$$P(S_T) = \prod_{t=1}^{T} P(e_t | e_1, e_2, \dots, e_{t-1})$$

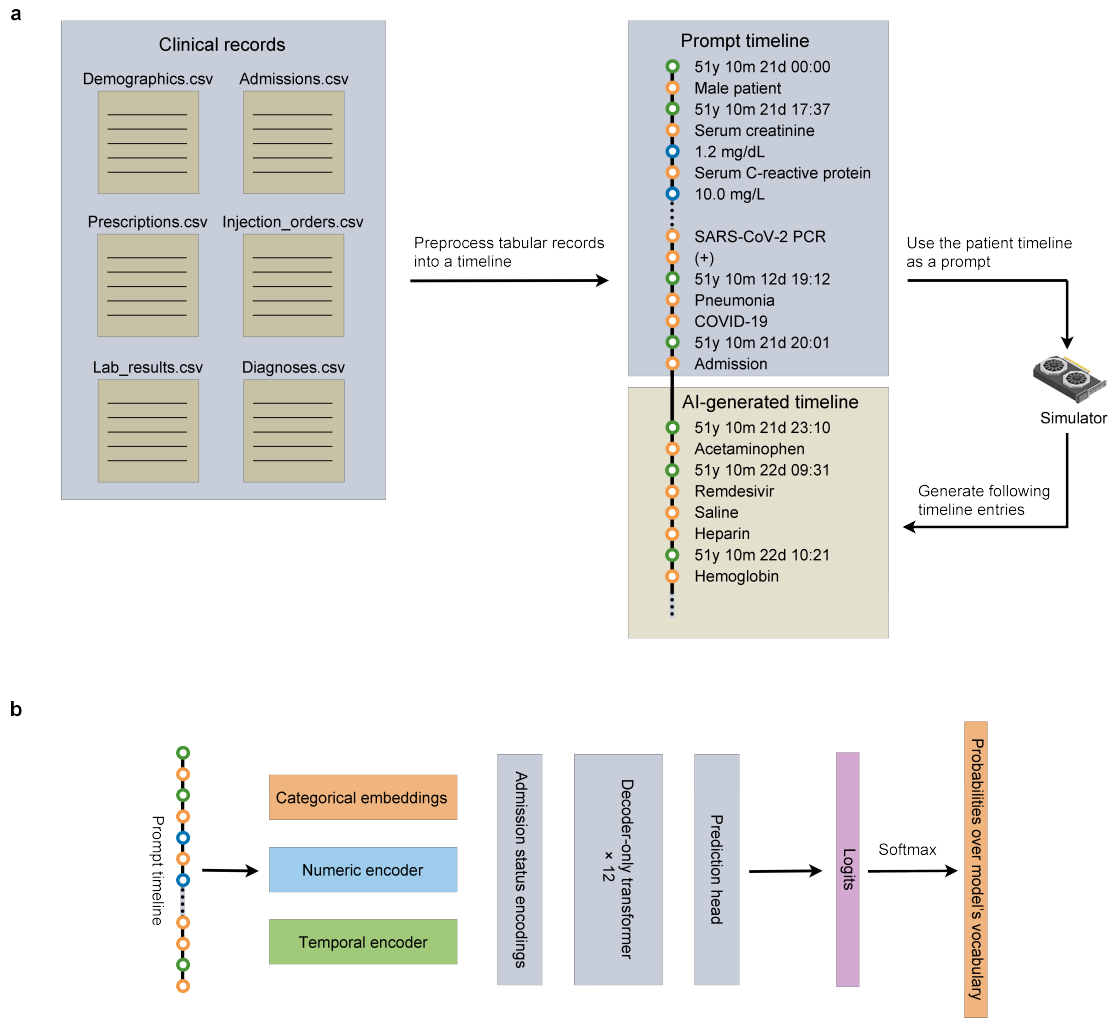

**Figure 1 | Overview of the simulation model. a**. Autoregressive decoding process. Structured clinical records were converted into temporally ordered patient timelines. Timeline entries were arranged sequentially after patient age with minute-level resolution. Laboratory test results and discharge outcomes were positioned immediately after the corresponding entries. Patient sex was represented as a token, and always appended at the beginning of the timeline. The model received the timeline as a prompt and autoregressively generated future timeline entries, one at a time until a stopping condition was met. **b**. Model architecture. Each timeline entry was encoded using a categorical, numeric, or temporal encoder and passed through 12 decoder-only transformer blocks. The final output was used to compute a probability distribution over the vocabulary from which the next entry is sampled.

, where each entry $e_i$ is generated sequentially, based on the history of prior entries.

**Model architecture**

The model architecture is shown in Figure 1. The simulation model used in this study builds on the architecture of our previously published autoregressive transformer model[5], with two key modifications. First, we introduced admission encoding. This is a learnable vector added to each timeline entry occurring during an admission period. Second, numeric laboratory test results were normalized using percentile-based scaling and discretized into 2,000 uniform bins spanning the 0th to 100th percentiles.

Each timeline entry (i.e., prompt) was encoded using one of three encoding pathways, selected based on the data type. Categorical entries—including diagnoses, medications, laboratory test codes, non-numeric lab results, admissions, discharges, and discharge outcomes (e.g., alive or deceased)—were encoded via embedding layers. Numeric values were processed through a dedicated numeric encoder comprising two feed-forward layers with GeLU activations[10]. Patient age was represented as a five-dimensional vector [years, months, days, hours, minutes], normalized to the [0,1] range, and passed through a temporal encoder with the same architecture as the numeric encoder.

Encoded vectors were interleaved in chronological order to construct the input matrix, which was passed through a stack of 12 decoder-only transformer blocks. Each block followed the GPT-2 architecture[11], including its weight initialization scheme. The final row of the transformer output was treated as a whole-timeline representation and fed into a prediction head to compute the next-entry probability distribution over the model's vocabulary. The model's hyperparameters are summarized in Table 1. We adapted the hyperparameters from the GPT-2 study to fit our research's hardware resources, customizing them to ensure that the model could be trained and tested within a practical time frame given our hardware capabilities.

The vocabulary was derived from the training dataset. Each unique categorical value was treated as a distinct vocabulary entry. Discretized numeric test results were incorporated using the percentile-based binning scheme described above. Time progression entries were also discretized. For intervals of less than 24 hours, the time progression was segmented into 10-minute bins. Intervals exceeding 24 hours were categorized into four ranges (1–30, 31–180, 181–360, and 361–1800 days), each further divided using progressively coarser bins (1-day, 10-day, 30-day, and 180-day intervals). Within each bin, time-of-day sub-segmentation is applied using 1-hour intervals (e.g., 00:00–00:59, 01:00–01:59, …, 23:00–23:59).

**Table 1 | Model hyperparameters.**

| $N$ | $d_{model}$ | $d_{ff}$ | $n_{heads}$ | $d_{head}$ | $d_{seq}$ | $P_{drop}$ |
|---|---|---|---|---|---|---|
| 12 | 768 | 3072 | 12 | 64 | 2048 | 0.1 |

\* $N$, number of transformer blocks; $d_{model}$, model dimension; $d_{ff}$, feed-forward hidden layer dimension; $n_{heads}$, number of attention heads; $d_{head}$, dimension of each attention head; $d_{seq}$, maximum sequence length (context window); $P_{drop}$, dropout rate.

**Model pretraining**

The model was trained to minimize cross-entropy loss using the AdamW optimizer[12] with a weight decay of 0.01. The learning rate was linearly warmed up over the first 1% of total training steps to a maximum of $5.0 \times 10^{-4}$, followed by cosine annealing to zero. The training was performed for 100 epochs using four NVIDIA A100 80GB GPUs. During the initial warmup phase, the batch size was gradually increased from 8 to a maximum of 32.

To enable unbiased evaluation of future data, only clinical records from January 2011 to December 2022 were used for model training. Records from 2023 were held out and exclusively used for counterfactual simulation.

**Counterfactual simulations of the COVID-19 admission cohort**

To assess the model's capacity for counterfactual reasoning, we conducted simulations on patients hospitalized with COVID-19 in 2023. We selected COVID-19 as the disease model due to its high prevalence and substantial mortality. It also spans diverse clinical contexts and is supported by extensive evidence generated during the global pandemic.[13]. We targeted three patient attributes for modification: age, serum C-reactive protein (CRP), and serum creatinine. These variables were chosen because they are routinely measured, readily available in electronic health records, and clinically associated with disease progression and treatment choices.

Admissions were included only if SARS-CoV-2 infection was confirmed by a positive PCR or antigen test using nasopharyngeal swab, sputum, or saliva samples 24 hours before or after the admission. All such admissions were used for age-based simulations. We selected a subset of patients with at least one recorded CRP value for CRP-based simulations. Similarly, creatinine-based simulations included only admissions with at least one post-admission serum creatinine measurement.

For age simulations, we added 5, 10, or 15 years to the recorded age. For CRP, we added 50, 100, 150, or 200 mg/L to the original value. For serum creatinine, we added 1, 3, or 5 mg/dL. Simulations were run for 7 days from the time of admission (age) or from the time of laboratory measurement (CRP and creatinine).

Event probabilities were estimated using a Monte Carlo method. Let $S$ denote the number of simulations, and $n_e$ the number of simulations in which a target event occurred. The estimated 7-day event probability $p_e$ was calculated as:

$$p_e = \frac{n_e}{S}$$

. We set $S = 256$ to ensure sufficient simulation size. We evaluated whether the event probabilities estimated by the counterfactual simulations aligned with expected effects based on established clinical evidence (Table 2).

Given the number of patients $N_p$ and the number of real events as $E_r$, we computed event rate in the real data ($R_r$) as:

$$R_r = \frac{E_r}{N_p}$$

We computed event rate in the simulation data ($R_s$) as:

$$R_s = \frac{\sum_{i=1}^{N_p} p_{ei}}{N_p}$$

. We then compared $R_r$ and $R_s$. In typical simulation setups, simulated results are compared with true outcomes. However, the true outcomes of counterfactual trajectories can never be observed, making a direct comparison unfeasible. Instead, we compared $R_r$ and $R_s$ to assess whether they displayed expected patterns (i.e., $R_s$ is expected to be higher than $R_r$, and indeed, $R_s$ was higher than $R_r$).

Simulations were performed using the same four GPUs employed during model training, with one model instance allocated per GPU. Prompts were placed in a shared queue and dynamically fetched by workers as they completed prior jobs, enabling efficient parallel execution. Key-value caching was implemented to further accelerate generation[14].

To assess statistical differences between counterfactual and actual conditions, we performed two-sided independent two-sample t-tests (Welch's t-test). The actual condition was used as the reference. A 95% percentile ranges were estimated using non-parametric bootstrapping with 1,000 iterations.

**Table 2 | Reference assumptions for validating counterfactual simulation outcomes**

| Modification | Rate of remdesivir use | In-hospital death rate | Rate of hospital stay > 7 days |
| --- | --- | --- | --- |
| Increased age | Increase[*1] | Increase[*1] | Increase[*1] |
| Increased serum CRP | Increase[*2] | Increase[*2] | Increase[*2] |
| Increased serum Cre | Decrease[*3] | Increase[*4] | Increase[*4] |

*1 Older age is a known risk factor for disease progression[13]. *2 Elevated CRP reflects inflammation and disease severity, and it is known to be associated with disease progression[15]. *3 Remdesivir was recommended to be used with caution in patients with impaired renal function[13]. *4 Reduced kidney function is associated with poor clinical outcomes[16,17]. CRP: C-reactive protein; Cre: creatinine.

## Results
### Dataset

We identified 373,890 unique patients with at least one clinical encounter recorded between January 2011 and December 2023. For general model evaluation purposes, we reserved the most recent 7% of patients (ordered by their first visit dates) as a holdout test set. The remaining patients were randomly split into training and validation sets, comprising 340,659 and 7,496 patients, respectively. The median age in the training set was 50 years, and 52.7% of patients were female. We created this dataset to train and validate our model for a variety of downstream applications, including the present study. Therefore, the data collection and split strategy are not tailored solely to this study, but are designed to be adaptable to different settings.

We selected COVID-19 admissions from 2023 regardless of their original data split. Although some patients may have appeared in the training dataset through earlier encounters, their 2023 admission records were not used during model training. The model was trained exclusively on data collected up to December 2022, ensuring that all 2023 COVID-19 admissions were temporally out-of-sample and used solely for counterfactual simulation. This design preserves a clear temporal boundary between training and evaluation[18].

### Model pretraining

The model pretraining was completed in 2 days and 5 hours. The final training and validation loss was 2.564 and 2.640, respectively. The model vocabulary included 17,995 unique items, composed of diagnosis codes (60.5%), numeric laboratory values (11.1%), laboratory test names (10.3%), medication codes (9.1%), time progression markers (8.7%), and other tokens (0.4%). The total number of trainable parameters in the model was 146,281,280.

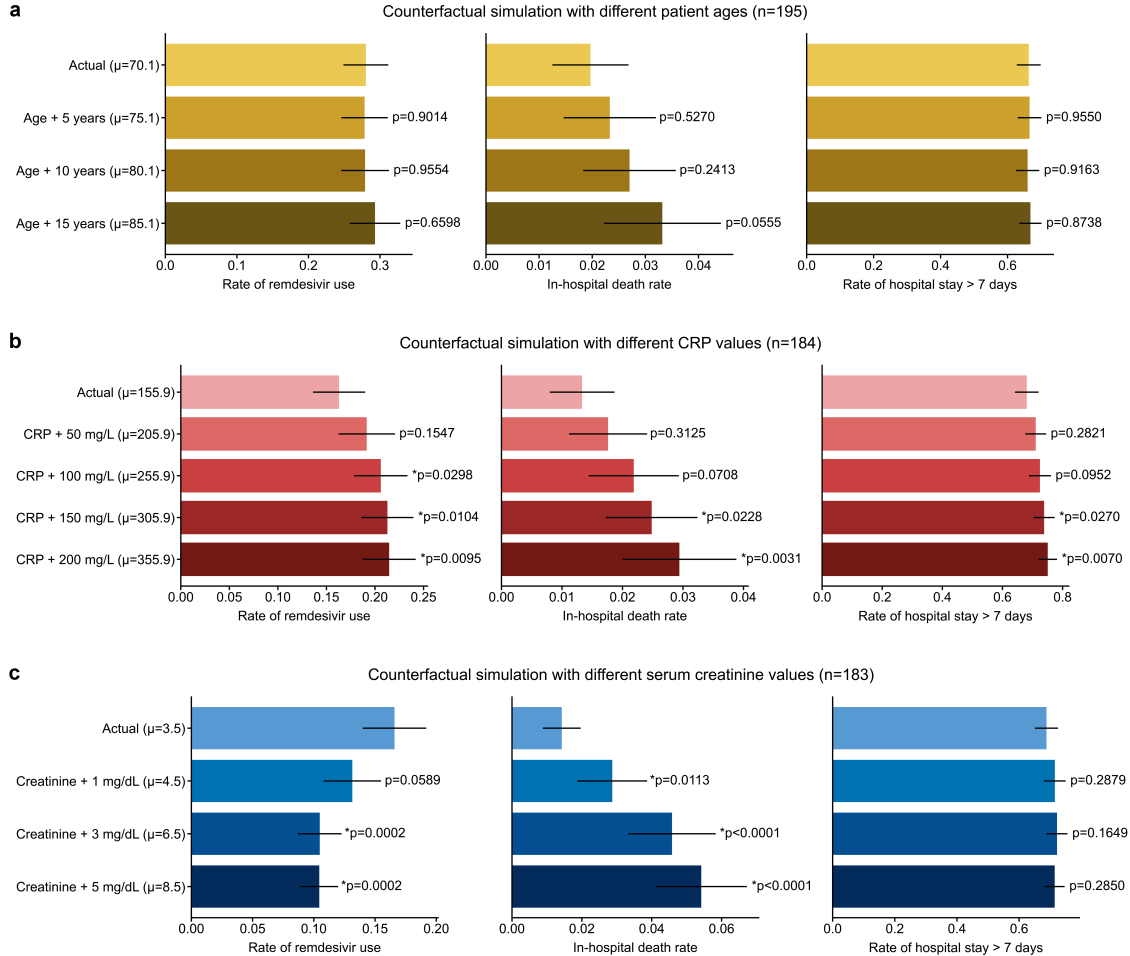

**Figure 2 | Counterfactual simulations aligned with expected clinical trends.** *p*-values are displayed alongside the bars; values ≤ 0.05 are marked as statistically significant. Error bars represent the 95th percentile range of the estimated probabilities. **a**. Simulations with increased patient age (n = 195). **b**. Simulations with elevated serum C-reactive protein (CRP) values (n = 184). All outcomes showed increasing trends with higher CRP. **c**. Simulations with increased serum creatinine values (n = 183). Remdesivir use decreased with higher creatinine, while mortality and prolonged hospitalization rates increased.

## Counterfactual simulation of age—what if the patient were older?

We identified 195 eligible COVID-19 admissions involving 191 unique patients (47.7% female). All admissions were used for counterfactual simulations in which patient age was modified. Each admission underwent 256 simulations over a 7-day period, yielding 4,864 simulated patient trajectories and 34,048 total patient days.

Simulation results are summarized in Figure 2a. In-hospital mortality increased with advancing simulated age, with the strongest association observed at age + 15 years ($p = 0.0055$). In contrast, rates of remdesivir use and prolonged hospitalization (>7 days) did not exhibit consistent trends across age-modified scenarios.

## Counterfactual simulation of serum CRP—what if inflammation were worse?

We excluded 11 admissions without recorded CRP values. Counterfactual simulations with increased CRP levels are summarized in Figure 2b. Remdesivir use increased with higher CRP levels ($p$-value range: 0.1547–0.0095), reflecting a tendency to treat patients with more apparent inflammatory responses. In-hospital death rates also increased with higher CRP ($p$-value range: 0.3125–0.0031). The probability of hospitalization longer than 7 days was significantly elevated when CRP was increased by 150 or 200 mg/L ($p = 0.0270$ and 0.0070, respectively).

**Counterfactual simulation of serum creatinine—what if kidney function were worse?**

We excluded 12 admissions without serum creatinine measurements. Simulation results for modified creatinine values are shown in Figure 2c. As expected, remdesivir use declined with impaired kidney function ($p$-value range: 0.0589–0.0002), consistent with its recommendations in patients with renal dysfunction during the study period. In-hospital death rates increased with worsening kidney function ($p$-value range: 0.0113 to <0.0001). Although not statistically significant, there was a trend toward longer hospital stays in simulations involving renal impairment ($p$-value range: 0.2879–0.1649).
A total of 622,592 counterfactual simulations—across age, CRP, and creatinine modification experiments—were completed in 9 hours and 3 minutes, yielding an average runtime of 0.052 seconds per simulation.

**Discussion**

In this study, we presented a generative deep-learning framework capable of simulating patient trajectories under counterfactual conditions using real-world clinical data. By modifying individual patient attributes—age and laboratory test results—we demonstrate that the model produces outcome shifts that align with known clinical patterns. Classic counterfactual simulation approaches based on non-generative machine learning methods, such as decision trees, are typically restricted to single-outcome prediction tasks[20]. Our model operates as a generative foundation model capable of producing patient timelines. Therefore, this framework enables flexible counterfactual simulations by allowing arbitrary modifications to any element of the input prompt, including laboratory values, diagnoses, medications, or time intervals. To our knowledge, this is the first generative model architecture designed to simulate patient timelines under a broad range of hypothetical scenarios. Our work presents a promising framework for future applications of autoregressive modeling in personalized medicine, in silico experimentation, and dynamic decision support.

A generative model like this can be valuable for various downstream applications. It can help increase the representation of specific patient data in clinical datasets. For example, death cases are often underrepresented in clinical data, but synthetic patient data can be generated by modifying prompts with attributes linked to higher mortality, as demonstrated in our study. While our research focused on COVID-19 patients, this approach can be extended to any medical condition. Exploring different scenarios could also enhance patient-physician interactions. For instance, counterfactual simulations showing higher mortality at older ages may encourage patients to start treatment sooner rather than delay it. In summary, this model offers a powerful tool for improving both data representation and clinical decision-making.

Counterfactual simulations involving CRP and creatinine values showed expected patterns. Elevated CRP, a marker of systemic inflammation, was associated with increased remdesivir use, higher mortality, and longer hospital stays. Likewise, increased serum creatinine, which indicates impaired kidney function, was linked to reduced remdesivir use and elevated risk of adverse outcomes. These findings align with established clinical knowledge and treatment practices, supporting the model's ability to reproduce real-world associations[13-15]. Notably, the model achieved this performance without access to explicit causal labels, outcome annotations, or rule-based constraints during training. Instead, it was trained using a self-supervised objective on a large-scale, longitudinal real-world dataset. The emergence of clinically meaningful patterns in the model's counterfactual outputs suggests that self-supervised generative learning is capable of capturing underlying physiological relationships and treatment behaviors embedded in real-world data. This underscores a previously underutilized potential of real-world data—when paired with large-scale generative models—to enable exploratory clinical simulations, offering new pathways for advancing evidence-based decision-making[20].

Simulations involving age modification, however, did not produce consistent changes in remdesivir use or length of stay, and only modest increases in mortality were observed at the highest age increment. Several factors may explain this. First, the actual patient cohort was already relatively old (mean age > 65 years old), potentially limiting the marginal effect of additional age[13]. Second, during the study period, COVID-19 admissions were subject to regulations that imposed discharge criteria, particularly for infection control purposes. Even patients with mild disease were often required to remain hospitalized for a predefined duration. These constraints may have reduced variation in hospitalization length and weakened the influence of individual-level factors such as age on discharge timing or treatment decisions. Accordingly, the absence of significant differences in some age-modified scenarios may not reflect a limitation of the model but rather suggest that it captured the limited marginal impact of age within this context.

While this study focused on COVID-19 admissions, the underlying model architecture is disease-agnostic. The model was trained on a vocabulary of over 10,000 clinical entries, covering more than 1,000 distinct medications and laboratory tests, as well as thousands of diagnostic codes. This broad coverage allows the model to adapt easily to a wide range of clinical settings, such as chronic disease management, pediatrics, and oncology. From a technical standpoint, counterfactual simulation across other disease domains can already be implemented with no modifications to the model. Moreover, we demonstrated that such simulations can typically be completed in less than one second when paired with modern GPU resources and key-value caching[14]. Together, these features position the model as a versatile and scalable tool for clinical simulation across diverse conditions.

This study has limitations. Counterfactual modifications were limited to a small set of variables and applied independently. The analysis focused exclusively on COVID-19 admissions. While evaluating alternative treatment strategies—such as medication choices—is a critical objective for counterfactual simulation in clinical medicine, we did not explore such interventions in this study due to unresolved methodological challenges. Simulating counterfactual treatment choices, in particular, requires robust causal simulation modeling. Furthermore, treatment-effect counterfactual prediction remains an emerging area, and widely accepted frameworks for evaluating such simulations are not yet established[20]. The validity of our modeling strategy for future downstream applications requires testing across more diverse scenarios, along with the development of evaluation frameworks for counterfactual predictions. We included patients from the training dataset for the simulation assessment due to the small number of events, which limited statistical power. However, we modified the real patient prompts to ensure that the model generated trajectories distinct from those it had encountered during training.

**Conclusion**
A generative autoregression decoder model trained with self-supervised learning successfully learned patient timeline dynamics embedded in real-world data. The model reproduced expected counterfactual consequences by using artificially modified patient timelines as prompts. Although further validation across more diverse clinical scenarios is necessary, this study represents an early step toward the clinical application of counterfactual simulation.

**Acknowledgments**
This study was supported by JST SPRING (Grant Number JPMJSP2108); Japan Agency for Medical Research and Development (AMED) "Moonshot Project Goal-7" under Grant Number JP22zf0127006; and the Cross-ministerial Strategic Innovation Promotion Program (SIP) on "Integrated Health Care System" (Grant Number JPJ012425). This work was supported by the Saito Scholarship of the Japanese Society for Medical and Biological Engineering.